\documentclass[twocolumn]{article}
\usepackage[utf8]{inputenc}
\usepackage{graphicx}
\usepackage{subcaption}
\usepackage{booktabs}
\usepackage{hyperref}
\usepackage{amsmath}
\usepackage{amsfonts}
\usepackage{amssymb}
\usepackage{xcolor}
\usepackage{longtable}
\usepackage{polyglossia}
\setdefaultlanguage{english}
\setotherlanguage{arabic}
\usepackage{url}

\usepackage[left=1in,right=1in,top=1in,bottom=1in]{geometry}
\usepackage{fancyhdr}
\usepackage{enumitem}

\fancypagestyle{firstpage}{
  \fancyhf{}

  \fancyhead[L]{\textbf{AtlasIA}}
  \fancyhead[R]{\textbf{2025-09-17}}
  \fancyfoot[C]{\thepage}
}
\title{\textbf{AtlasOCR: Building the First Open-Source Darija OCR Model with Vision Language Models}}

\author{
  Imane Momayiz, Soufiane Ait Elaouad, Abdeljalil Elmajjodi, Haitame Bouanane \\
  \texttt{https://www.atlasia.ma/} \\
  \texttt{https://github.com/atlasia-ma/}
}

\date{} 

\begin{document}

\maketitle
\thispagestyle{firstpage}

\begin{abstract}
Darija, the Moroccan Arabic dialect, is rich in visual content yet lacks specialized Optical Character Recognition (OCR) tools. This paper introduces \textbf{AtlasOCR}, the first open-source Darija OCR model built by fine-tuning a 3B parameter Vision Language Model (VLM). We detail our comprehensive approach, from curating a unique Darija-specific dataset leveraging both synthetic generation with our OCRSmith library and carefully sourced real-world data, to implementing efficient fine-tuning strategies. We utilize QLoRA and Unsloth for parameter-efficient training of Qwen2.5-VL 3B and present comprehensive ablation studies optimizing key hyperparameters. Our evaluation on the newly curated AtlasOCRBench and the established KITAB-Bench demonstrates state-of-the-art performance, challenging larger models and highlighting AtlasOCR's robustness and generalization capabilities for both Darija and standard Arabic OCR tasks.
\end{abstract}

\section{Introduction}

Darija, the Moroccan Arabic dialect, represents a vibrant and visually rich linguistic landscape prevalent in social media, informal documents, and handwritten materials. Despite its widespread use and unique characteristics, the absence of dedicated Optical Character Recognition (OCR) tools for Darija has posed significant barriers for developers, researchers, and organizations working with Moroccan content. This gap hinders digital preservation efforts, limits large-scale text analysis capabilities, and impedes accessibility initiatives.

To address this critical need, we introduce \textbf{AtlasOCR}, the first open-source OCR model specifically designed for Darija. AtlasOCR is a 3-billion-parameter model developed by fine-tuning a Vision Language Model (VLM) using parameter-efficient techniques. This paper presents our comprehensive methodology, spanning data curation, model selection, training strategies, and extensive evaluation.

\textbf{Our key contributions include:}
\begin{itemize}[noitemsep,topsep=0pt]
    \item Development of \textbf{AtlasOCR}, the first open-source Darija OCR model achieving state-of-the-art performance
    \item A comprehensive data curation strategy combining synthetic data generation via OCRSmith with diverse real-world Darija content
    \item Detailed methodology for parameter-efficient fine-tuning using QLoRA and Unsloth frameworks
    \item Extensive ablation studies optimizing key hyperparameters and training configurations
    \item Creation of \texttt{AtlasOCRBench}, a publicly available benchmark for Darija OCR evaluation
    \item Thorough evaluation demonstrating superior Darija performance and competitive results on standard Arabic benchmarks
\end{itemize}

\section{Related Work}

\subsection{Arabic OCR Systems}
Traditional Arabic OCR systems have primarily focused on Modern Standard Arabic (MSA), with limited attention to dialectal variants. Recent advances in deep learning have improved Arabic text recognition capabilities \cite{kitabbench}, but dialectal Arabic, particularly Darija, remains underexplored due to data scarcity and linguistic complexity.

\subsection{Vision Language Models for OCR}
Vision Language Models have revolutionized document understanding by integrating visual comprehension with linguistic context \cite{nanovlm}. These models excel at zero-shot generalization and have shown promising results in multilingual OCR tasks, making them ideal candidates for low-resource language applications.

\section{Background}

\subsection{The Importance of Darija OCR}

The development of Darija OCR capabilities offers substantial benefits across multiple domains:

\begin{itemize}[noitemsep,topsep=0pt]
    \item \textbf{Digital Preservation}: Converting historical documents, manuscripts, and traditional texts into searchable digital formats
    \item \textbf{Social Media Analysis}: Enabling large-scale analysis of public discourse and sentiment in Moroccan online content
    \item \textbf{Accessibility}: Making visual content accessible to screen readers and assistive technologies
    \item \textbf{Research Applications}: Facilitating linguistic studies, cultural heritage research, and corpus development
\end{itemize}

\subsection{Vision Language Model Architecture}

\begin{figure}[h]
    \centering
    \includegraphics[width=0.48\textwidth]{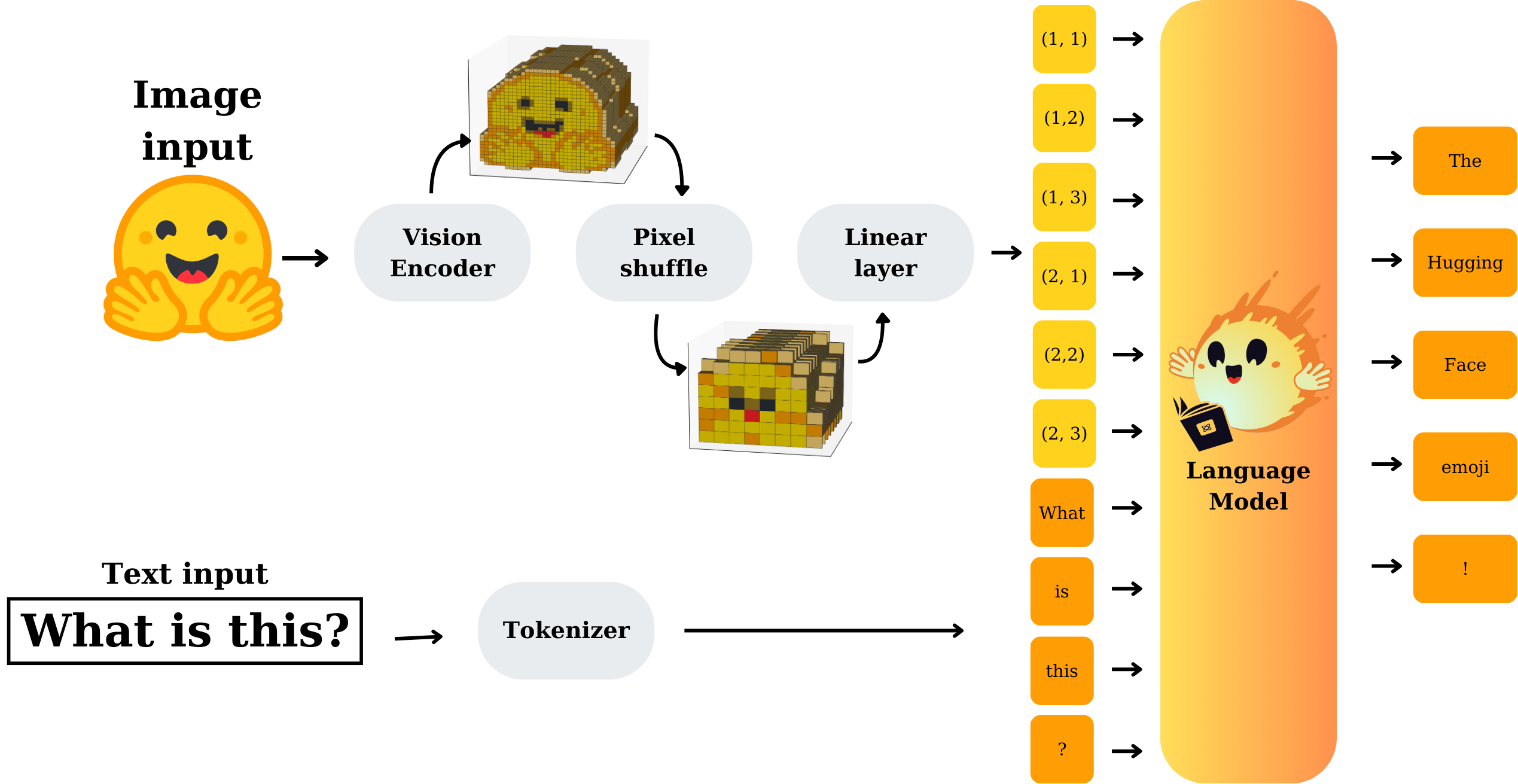}
    \caption{Vision Language Model Architecture \cite{nanovlm}}
    \label{fig:vlm_architecture}
\end{figure}

Vision Language Models integrate three core components to process multimodal inputs (Figure \ref{fig:vlm_architecture}):

\begin{itemize}[noitemsep,topsep=0pt]
    \item \textbf{Vision Encoder}: Converts input images into high-dimensional vector embeddings capturing visual features such as colors, shapes, textures, and spatial arrangements
    \item \textbf{Modality Projection Module}: Aligns visual features with the language model's representation space, ensuring semantic consistency
    \item \textbf{Language Model}: Integrates aligned visual embeddings with textual prompts to generate coherent natural language outputs
\end{itemize}

For OCR applications, this architecture enables understanding both visual text layout and linguistic nuances, crucial for accurately recognizing Darija text across diverse fonts, styles, and backgrounds.

\section{Data Curation}

Developing a robust Darija OCR model required creating a large-scale, diverse dataset reflecting real-world variability. Our hybrid approach combined synthetic data generation with carefully curated real-world samples.

\subsection{Synthetic Data Generation with OCRSmith}

Given the high cost and time requirements for annotating quality datasets in under-resourced languages, synthetic data generation provided an efficient scaling solution. We developed \texttt{OCRSmith} \cite{ocrsmith}, an open-source toolkit that simulates realistic text conditions including various fonts, layouts, backgrounds, and distortions.

\begin{figure}[h]
    \centering
    \includegraphics[width=0.48\textwidth]{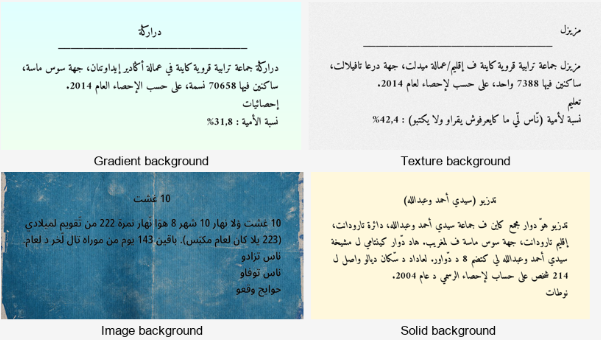}
    \caption{Synthetic Darija Text Examples Generated with OCRSmith}
    \label{fig:ocrsmith_examples}
\end{figure}

OCRSmith enables rapid generation of thousands of labeled images complete with bounding boxes and metadata. Figure \ref{fig:ocrsmith_examples} demonstrates synthetic Darija text examples generated using this toolkit.

\subsection{Real-World Data Collection}

While synthetic data provided scale, real-world images ensured authenticity and captured nuances difficult to simulate. We curated diverse Darija text from multiple contexts:

\begin{itemize}[noitemsep,topsep=0pt]
    \item \textbf{Scanned Literature}: Two key sources provided approximately 700 pages of high-quality Darija text: "العَرَبِيَّةُ الدَّارِجَةُ" by Mohammed El-Madlaoui El-Mounabhi and "علشان الصغيرة والصغير" by Farouk ElMarrakchi. These were pseudo-labeled using Gemini 2.0 Flash.
    
    \item \textbf{Social Media Content}: LinkedIn and similar platforms yielded poster-style educational materials converted to images suitable for OCR training.
    
    \item \textbf{Educational Materials}: Moroccan study materials, particularly driving license exam preparations, provided challenging samples with varied text quality and layouts.
    
    \item \textbf{Recipe Collections}: Moroccan cookbooks offered domain-specific vocabulary and formatting, carefully preprocessed for optimal OCR training.
\end{itemize}

\begin{figure}[h]
    \centering
    \begin{subfigure}[b]{0.48\textwidth}
        \centering
        \includegraphics[width=\textwidth]{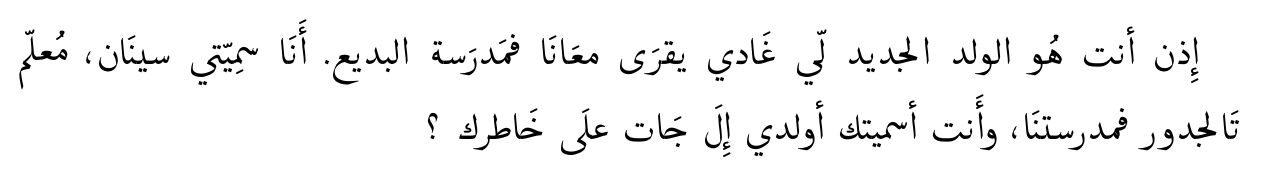}
        \caption{Scanned Literature}
        \label{fig:scanned_book}
    \end{subfigure}
    \hfill
    \begin{subfigure}[b]{0.48\textwidth}
        \centering
        \includegraphics[width=\textwidth]{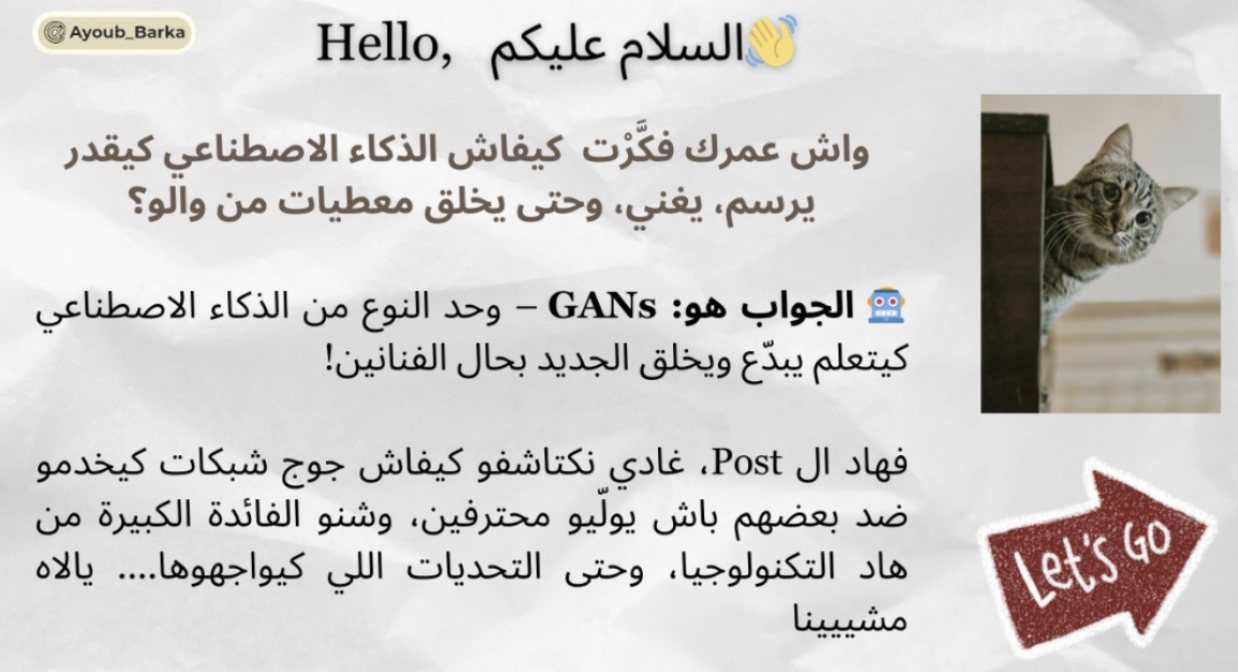}
        \caption{Social Media Content}
        \label{fig:social_media}
    \end{subfigure}
    
    \begin{subfigure}[b]{0.48\textwidth}
        \centering
        \includegraphics[width=\textwidth]{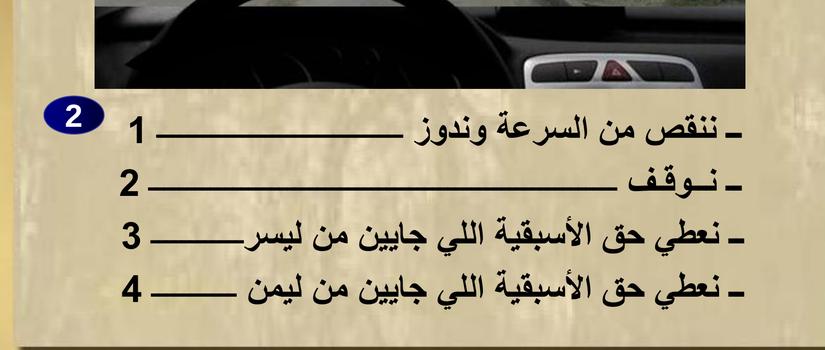}
        \caption{Educational Materials}
        \label{fig:permis}
    \end{subfigure}
    \hfill
    \begin{subfigure}[b]{0.48\textwidth}
        \centering
        \includegraphics[width=\textwidth]{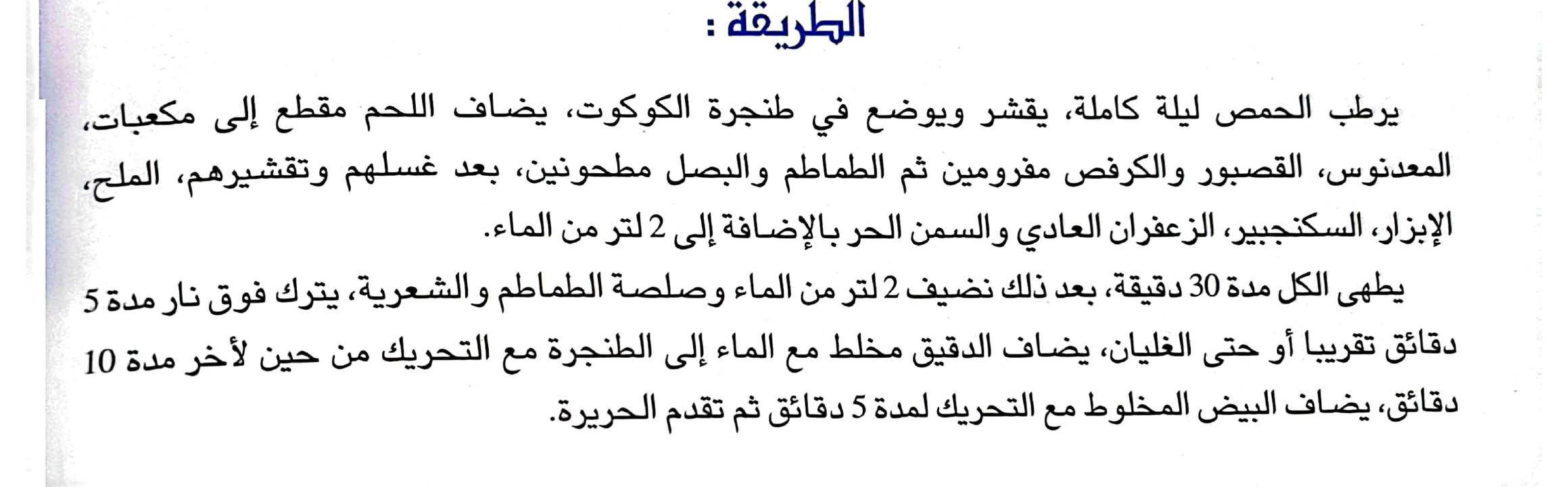}
        \caption{Recipe Collection}
        \label{fig:cookbook}
    \end{subfigure}
    \caption{Real-World Darija Text Sources}
    \label{fig:real_world_data}
\end{figure}

Figure \ref{fig:real_world_data} illustrates examples from each real-world data source.

\subsection{Dataset Composition}

Our hybrid approach yielded the first large-scale Darija OCR dataset. Table \ref{tab:dataset_overview} summarizes the dataset composition, with approximately 86\% synthetic and 14\% real-world content, ensuring both the scale necessary for robust training and the authenticity required for real-world applicability.

\begin{table}[h]
    \centering
    \caption{Darija OCR Dataset Overview}
    \label{tab:dataset_overview}
    \begin{tabular}{lcc}
        \toprule
        \textbf{Split} & \textbf{Samples} & \textbf{Total Words} \\
        \midrule
        Train & 26,162 & 9.5M \\
        Validation & 3,930 & 1.2M \\
        \textbf{Total} & \textbf{30,092} & \textbf{10.7M} \\
        \bottomrule
    \end{tabular}
\end{table}

\section{Methodology}

\subsection{Base Model Selection}

Selecting an appropriate base model was critical for developing high-performance Darija OCR within computational constraints. We benchmarked several open-source vision-language models on 55 manually curated real-world Darija images:

\begin{itemize}[noitemsep,topsep=0pt]
    \item Qwen2-VL 2B \cite{qwen2vl2b}
    \item Qwen2.5-VL 3B \cite{qwen25vl3b}
    \item Qari OCR 2B \cite{qariocr2b}
    \item ArabicNougat \cite{arabicnougat}
\end{itemize}

We prioritized compact architectures (2B-3B parameters) for efficient training and inference. Evaluation results consistently showed \textbf{Qwen2.5-VL 3B} outperforming alternatives across diverse Darija text domains, establishing it as our base model.

\subsection{Parameter-Efficient Fine-Tuning Strategy}

We employed a parameter-efficient approach combining Quantized Low-Rank Adaptation (QLoRA) with Unsloth optimization:

\begin{itemize}[noitemsep,topsep=0pt]
    \item \textbf{QLoRA} \cite{qlora}: Enables efficient fine-tuning by quantizing models to 4-bit precision with low-rank adapters, reducing memory requirements by up to 80\% while maintaining performance.
    
    \item \textbf{Unsloth} \cite{unsloth}: Accelerates LLM fine-tuning with up to 5× speed improvements and 60\% memory reduction through optimized GPU kernels and memory management.
\end{itemize}

This combination enabled effective 3B parameter model fine-tuning on consumer-grade GPUs, ensuring accessibility and resource efficiency.

\section{Experimental Design}

\subsection{Ablation Studies}

We conducted comprehensive ablation studies to optimize AtlasOCR's performance, focusing on key hyperparameters and training configurations.

\subsubsection{LoRA Hyperparameters}

We explored combinations of LoRA rank ($r$), scaling factor ($\alpha$), and dropout rate, measuring impact via minimum evaluation loss (Table \ref{tab:lora_ablation}).

\begin{table}[h]
    \centering
    \caption{LoRA Hyperparameter Ablation Results}
    \label{tab:lora_ablation}
    \begin{tabular}{@{}cccc@{}}
        \toprule
        \textbf{$r$} & \textbf{$\alpha$} & \textbf{Dropout} & \textbf{Eval Loss} \\
        \midrule
        32 & 32 & 0.00 & 0.2442 \\
        32 & 32 & 0.05 & 0.2456 \\
        64 & 64 & 0.05 & 0.2251 \\
        128 & 128 & 0.05 & \textbf{0.2132} \\
        \bottomrule
    \end{tabular}
\end{table}

Results indicate that increasing both rank and scaling factor improves performance, suggesting that additional adapter parameters enhance task-specific feature learning. The optimal configuration was $r=128$, $\alpha=128$, dropout=0.05.

\subsubsection{Quantization Precision Impact}

We compared 4-bit and 16-bit precision effects during fine-tuning (Table \ref{tab:quantization_ablation}).

\begin{table}[h]
    \centering
    \caption{Quantization Precision Comparison}
    \label{tab:quantization_ablation}
    \begin{tabular}{lc}
        \toprule
        \textbf{Precision} & \textbf{Min Eval Loss} \\
        \midrule
        4-bit & \textbf{0.2132} \\
        16-bit & 0.2124 \\
        \bottomrule
    \end{tabular}
\end{table}

The minimal performance difference validates 4-bit quantization for our task, offering substantial memory and computational savings without compromising accuracy.

\subsubsection{Optimization Parameters}

We investigated optimal batch size and learning rate combinations with different LoRA configurations (Tables \ref{tab:batch_lr_16} and \ref{tab:batch_lr_128}).

\begin{table}[h]
    \centering
    \caption{Optimization Results (LoRA $r=\alpha=16$)}
    \label{tab:batch_lr_16}
    \begin{tabular}{ccc}
        \toprule
        \textbf{Batch Size} & \textbf{Learning Rate} & \textbf{Min Eval Loss} \\
        \midrule
        16 & 2e-4 & 0.3161 \\
        16 & 6e-4 & \textbf{0.2326} \\
        16 & 2e-3 & 4.0958 \\
        64 & 2e-4 & 0.3479 \\
        128 & 2e-4 & 0.4086 \\
        128 & 8e-4 & 0.2733 \\
        128 & 2e-3 & 0.2350 \\
        512 & 2e-4 & 0.5725 \\
        \bottomrule
    \end{tabular}
\end{table}

\begin{table}[h]
    \centering
    \caption{Optimization Results (LoRA $r=\alpha=128$)}
    \label{tab:batch_lr_128}
    \begin{tabular}{ccc}
        \toprule
        \textbf{Batch Size} & \textbf{Learning Rate} & \textbf{Min Eval Loss} \\
        \midrule
        16 & 6e-5 & 0.2652 \\
        16 & 2e-4 & \textbf{0.2132} \\
        128 & 2e-4 & 0.2456 \\
        128 & 8e-4 & 0.2165 \\
        128 & 2e-3 & 8.2561 \\
        \bottomrule
    \end{tabular}
\end{table}

Higher learning rates generally improved convergence speed when not causing instability. Increasing LoRA parameters necessitated learning rate recalibration for optimal performance.

\subsubsection{Vision Layer Fine-Tuning}

We evaluated the impact of freezing vision layers during training (Table \ref{tab:vision_freeze}).

\begin{table}[h]
    \centering
    \caption{Vision Layer Fine-Tuning Impact}
    \label{tab:vision_freeze}
    \begin{tabular}{lc}
        \toprule
        \textbf{Fine-tune Vision Layers} & \textbf{Min Eval Loss} \\
        \midrule
        Yes & \textbf{0.2155} \\
        No & 0.3173 \\
        \bottomrule
    \end{tabular}
\end{table}

Fine-tuning vision layers significantly improved performance, confirming the benefit of adapting these components to Darija's visual characteristics.

\subsubsection{RSLoRA Analysis}

We tested Rank-Stabilized LoRA (RSLoRA) \cite{rslora}, designed to improve scaling behavior with increasing adapter rank (Table \ref{tab:rslora_ablation}).

\begin{table}[h]
    \centering
    \caption{RSLoRA Impact Assessment}
    \label{tab:rslora_ablation}
    \begin{tabular}{lc}
        \toprule
        \textbf{RSLoRA Enabled} & \textbf{Min Eval Loss} \\
        \midrule
        No & \textbf{0.2132} \\
        Yes & 8.2561 \\
        \bottomrule
    \end{tabular}
\end{table}

RSLoRA significantly degraded performance in our setting, indicating it may not be suitable for all model architectures or tasks.

\section{Evaluation Framework}

\subsection{Benchmark Development}

We created \texttt{AtlasOCRBench}, a comprehensive evaluation benchmark tailored for Darija, integrating:

\begin{itemize}[noitemsep,topsep=0pt]
    \item \textbf{Scanned Darija Literature}: High-quality printed text representing foundational document types
    \item \textbf{OCRSmith Synthetic Data}: Controlled samples testing specific OCR challenges
\end{itemize}

Our benchmark creation pipeline (Figure \ref{fig:benchmark_pipeline}) employed a two-step process:

\begin{figure}[h]
    \centering
    \includegraphics[width=0.48\textwidth]{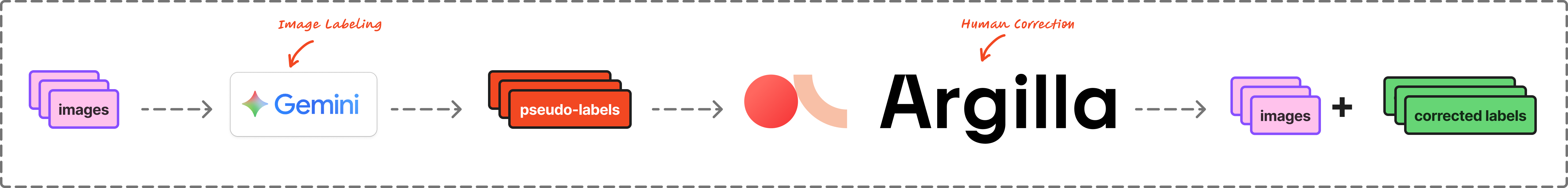}
    \caption{Benchmark Creation Pipeline}
    \label{fig:benchmark_pipeline}
\end{figure}

\begin{enumerate}[noitemsep,topsep=0pt]
    \item \textbf{Pseudo-labeling}: Using Gemini 2.0 Flash with carefully engineered prompts prioritizing human readability
    \item \textbf{Human Annotation}: Manual review and correction using Argilla for collaborative editing
\end{enumerate}

The final \texttt{AtlasOCRBench} contains 251 samples, including 55 from scanned literature, ensuring comprehensive coverage of realistic Darija OCR challenges.

\subsection{Evaluation Metrics}

We employed standard OCR evaluation metrics:

\begin{itemize}[noitemsep,topsep=0pt]
    \item \textbf{Character Error Rate (CER)}: Measures character-level editing distance normalized by ground truth length. Particularly suitable for Darija due to spelling variations and lack of standardized orthography.
    
    \item \textbf{Word Error Rate (WER)}: Measures word-level editing distance normalized by ground truth word count. While useful, can be misleading for Darija where single character differences mark entire words as incorrect.
\end{itemize}

We prioritize \textbf{CER as our primary metric} given Darija's linguistic characteristics and flexible spelling conventions.

\subsection{Preprocessing Protocol}

Our evaluation ensures fairness through consistent preprocessing:

\begin{enumerate}[noitemsep,topsep=0pt]
    \item \textbf{Text Normalization}:
    \begin{itemize}[noitemsep,topsep=0pt]
        \item Arabic diacritic removal (harakat)
        \item Line break standardization and whitespace normalization
    \end{itemize}
    \item \textbf{Metric Calculation}:
    \begin{itemize}[noitemsep,topsep=0pt]
        \item CER: Space-removed character-level comparison
        \item WER: Space-tokenized word-level comparison
    \end{itemize}
\end{enumerate}

\section{Results and Analysis}

\subsection{Benchmark Performance}

We evaluated AtlasOCR on both KITAB-Bench \cite{kitabbench} (a large-scale Arabic OCR benchmark with 8,800+ samples) and our AtlasOCRBench, providing comprehensive assessment across standard Arabic and Darija tasks.

\begin{figure}[h]
    \centering
    \includegraphics[width=0.48\textwidth]{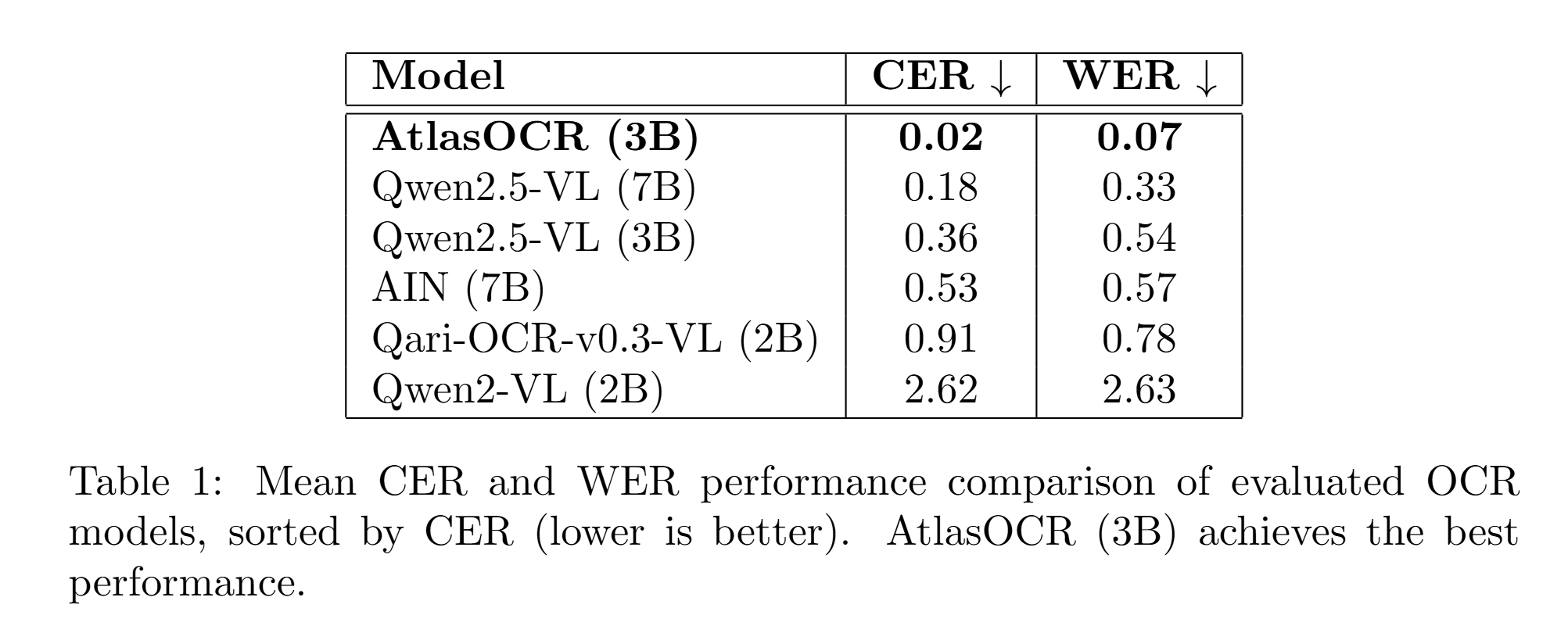}
    \caption{AtlasOCRBench Results (Lower CER indicates better performance)}
    \label{fig:atlasocrbench_leaderboard}
\end{figure}

\begin{figure}[h]
    \centering
    \includegraphics[width=0.48\textwidth]{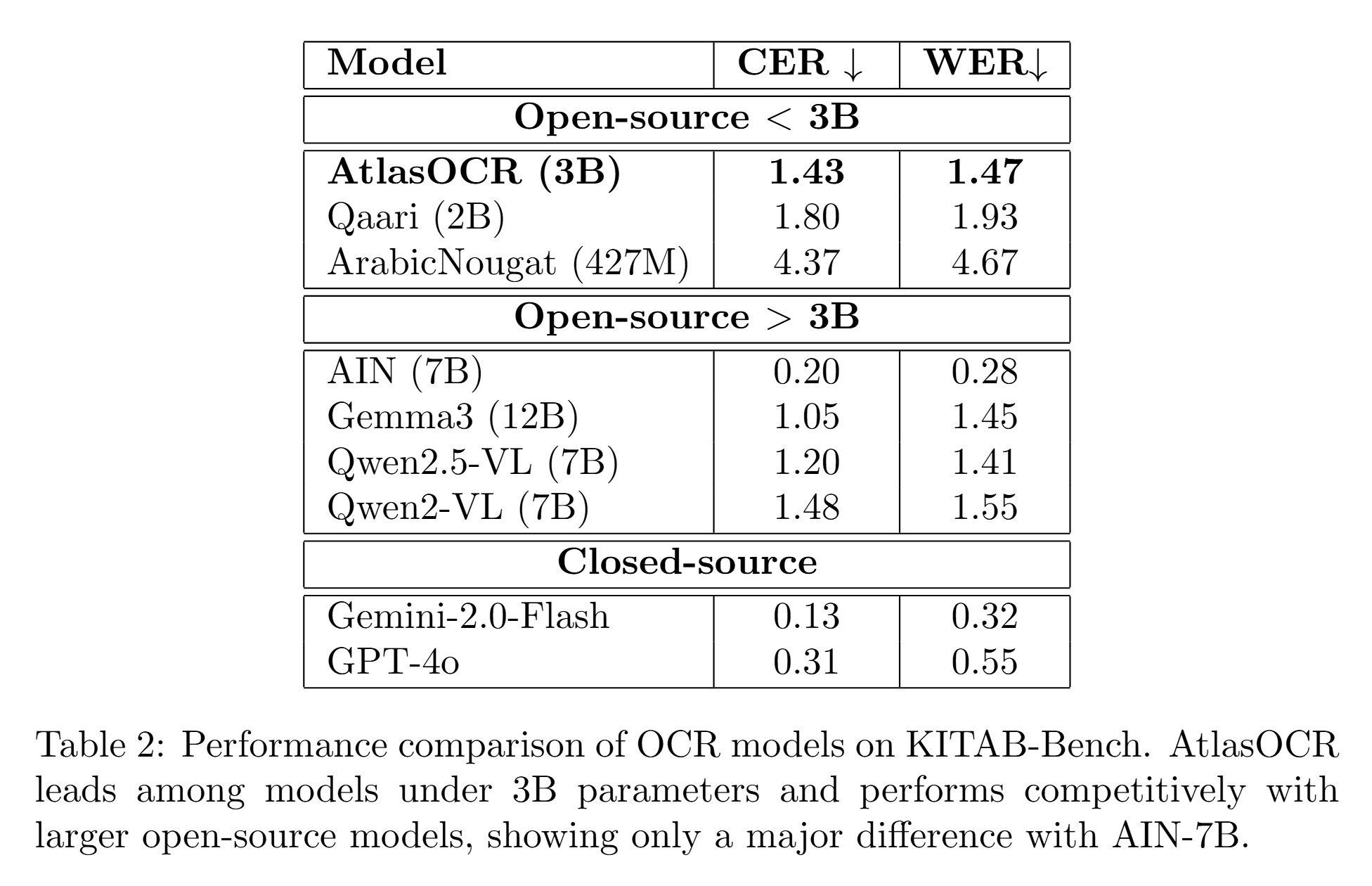}
    \caption{KITAB-Bench Results (Lower CER indicates better performance)}
    \label{fig:kitab_bench_leaderboard}
\end{figure}

Figure \ref{fig:atlasocrbench_leaderboard} demonstrates AtlasOCR's state-of-the-art performance on Darija OCR, significantly outperforming all open-source alternatives. Figure \ref{fig:kitab_bench_leaderboard} shows competitive performance on standard Arabic tasks, challenging much larger models including Gemma3 (12B) \cite{gemma} and Qwen2.5-VL (7B).

\subsection{Comparative Analysis}

AtlasOCR demonstrates superior performance on Darija-specific tasks while maintaining competitive results on standard Arabic benchmarks. The model achieves state-of-the-art results on AtlasOCRBench, significantly outperforming existing open-source alternatives. On KITAB-Bench, AtlasOCR competes effectively with much larger models, highlighting the effectiveness of our parameter-efficient fine-tuning approach.

Key performance insights include:
\begin{itemize}[noitemsep,topsep=0pt]
    \item \textbf{Darija Specialization}: AtlasOCR achieves the lowest CER on Darija text, validating our specialized training approach
    \item \textbf{Parameter Efficiency}: Competitive performance with significantly fewer parameters than larger alternatives
    \item \textbf{Cross-lingual Transfer}: Strong generalization to standard Arabic despite Darija-focused training
\end{itemize}

\section{Discussion}

\subsection{Key Findings}

Our experimental results reveal several important insights:

\begin{itemize}[noitemsep,topsep=0pt]
    \item \textbf{Parameter Efficiency}: AtlasOCR achieves superior performance with fewer parameters than competing models, demonstrating the effectiveness of our fine-tuning strategy
    \item \textbf{Cross-lingual Generalization}: Strong performance on standard Arabic benchmarks indicates robust generalization capabilities
    \item \textbf{Data Strategy Validation}: Our hybrid synthetic-real approach proved highly effective for low-resource language OCR development
\end{itemize}

\subsection{Limitations and Challenges}

Despite strong performance, AtlasOCR has certain limitations:

\begin{itemize}[noitemsep,topsep=0pt]
    \item \textbf{Diacritic Handling}: Limited capability for recognizing and reconstructing Arabic diacritics when present
    \item \textbf{Complex Layout Processing}: Performance may degrade on highly complex or artistic document structures
    \item \textbf{Domain Specificity}: Training data bias toward specific Darija domains may affect performance on underrepresented text types
\end{itemize}

\section{Conclusion and Future Directions}

AtlasOCR represents a significant advancement in Darija OCR, providing the first open-source solution addressing a critical gap for Moroccan content processing. Through careful dataset curation, efficient fine-tuning strategies, and comprehensive evaluation, we demonstrate that high-performance OCR models can be developed for under-resourced languages within reasonable computational constraints.

\subsection{Future Research Directions}

Our ongoing work focuses on:

\begin{itemize}[noitemsep,topsep=0pt]
    \item \textbf{Dataset Enhancement}: Expanding coverage of handwritten text, diacritized content, and diverse document types
    \item \textbf{Model Compression}: Developing sub-3B parameter variants for mobile and edge deployment
    \item \textbf{Layout Understanding}: Enhancing capabilities for complex document structures and mixed content
    \item \textbf{Multilingual Extension}: Adapting our methodology for other North African Arabic dialects
\end{itemize}

\subsection{Broader Impact}

AtlasOCR's development demonstrates the feasibility of creating high-quality language technology for under-resourced languages. Our open-source approach and detailed methodology provide a blueprint for similar initiatives, potentially catalyzing development of specialized tools for other dialectal variants and low-resource languages.

\section*{Acknowledgments}

We thank the Moroccan Arabic language community for their support in data collection and validation. Special recognition goes to contributors who helped identify and digitize historical Darija texts, making this work possible.

\bibliographystyle{plain}

\end{document}